# Towards socially-competent and culturally-adaptive artificial agents
## Expressive order, interactional disruptions and recovery strategies

Chiara Bassetti, Enrico Blanzieri, Stefano Borgo, Sofia Marangon

**Abstract**: The development of artificial agents for social interaction pushes to enrich robots with social skills and knowledge about (local) social norms. One possibility is to distinguish the expressive and the functional orders during a human-robot interaction. The overarching aim of this work is to set a framework to make the artificial agent socially-competent beyond dyadic interaction —interaction in varying multi-party social situations— and beyond individual-based user personalization, thereby enlarging the current conception of "culturally-adaptive". The core idea is to provide the artificial agent with the capability to handle different kinds of interactional disruptions, and associated recovery strategies, in microsociology. The result is obtained by classifying functional and social disruptions, and by investigating the requirements a robot's architecture should satisfy to exploit such knowledge. The paper also highlights how this level of competence is achieved by focusing on just three dimensions: (i) social capability, (ii) relational role, and (iii) proximity, leaving aside the further complexity of full-fledged human-human interactions. Without going into technical aspects, End-to-end Data-driven Architectures and Modular Architectures are discussed to evaluate the degree to which they can exploit this new set of social and cultural knowledge. Finally, a list of general requirements for such agents is proposed.

> Any serious effort to contend with the real time production and understanding of human actions in everyday interaction can scarcely avoid noting that they are characterized by the routine occurrence of troubles, "hitches," misunderstandings, "errors," and other infelicities.
>
> Raymond et al., 2013: 1

# 1. Introduction

Social action in interaction is a multi-layered phenomenon, complex in many respects. This is one, if not the, main challenge to human-robot interaction (HRI) and affects artificial intelligence (AI) more broadly. Multimodality of communication constitutes a complexity factor per se; yet it is fundamental for action coordination, mutual understanding, as much as the affective and emotional dimension of interpersonal communication. "Social robots must be able to read the different social and conversational cues that people use during interaction with each other, then use these to adapt [...] So it is imperative to investigate what types of [verbal and non-verbal] cues people use in their interaction" (Onyeulo & Gandhi, 2020: 5). This is an endeavor which, per se, has been extensively pursued in several disciplines, and prominently by Ethnomethodology and Conversation Analysis (EM/CA - cf., e.g., Garfinkel, 1967; Sacks, 1972), whose contribution could be further leveraged (as recognized in Human-Computer Interaction, e.g., Moore and Arar, 2019).

There is, furthermore, another layer of human action-in-interaction which is crucial and has to be addressed if we aim at developing socially-competent artificial agents. It is the expressive dimension, or the *expressive order* of social interaction (cf. Goffman, e.g., 1959, 1967). It has to do with social roles and positions, with participants' "faces" (impression management, deference and demeanor, reputation and respect), and with their social relationships (e.g., tacitly displaying that two participants hold a closer mutual relationship than with the other interactants). This is what has been called "interaction ritual" (Collins, 2004). The *expressive order is always operative when humans find themselves in the presence of others, whatever the activity* at stake (e.g., work or leisure) and the given scenario, or social occasion.

Such a social context does play an important role. For instance, Goffman (1961) metaphorically maintained that social encounters are characterized by an enveloping "membrane" which keeps outside "issues" considered irrelevant for the current social occasion. For example, in Western culture at a party at a friend's place, social class or salary are to be filtered out; in some societies, the same holds for gender issues at a job meeting or interview. Breaches in the membrane threaten the scenario itself, possibly causing a shift in the "frame" of activities (Goffman, 1974). To keep with the examples, a conversation and hence a social situation[1] suddenly stops to be a friendly one, a job interview may turn into a discriminatory encounter.

What is to be filtered out, furthermore, does not only change with the kind of encounter (social occasion/scenario), but also with changing culture (see, e.g., the gender example). Cultural variations —themselves multilayered and intersectional (national and regional cultures; professional or religious cultures, both possibly transnational yet with their local variety; occupational and organizational cultures; so-called "subcultures" of all sorts; etc.)— are orthogonal both to the functional and the expressive dimensions of human action in interaction. That is, the way things are done in a given *cultural context and social occasion/scenario* —i.e., the ethnomethods (Garfinkel, 1967, 2002; cf. also Liberman, 2013)— vary, and this is true for the functional as much as for the expressive order of interaction.

When considering human-robot interaction, therefore, culture is not only to be addressed in terms of its "influence on the acceptance [...] towards social robots" (Onyeulo & Gandhi, 2020: 4) or "on expectations towards and responses to social robots" (Lim et al., 2021: 307) —that is, as an individual phenomenon. When it comes to designing socially-competent artificial agents, equipping them with the ability to adapt to the local culture does not only nor primarily mean adaptation to a user's individual traits, including their cultural ones (i.e., adapting to a *kind* of individual[2]), but adaptation to different social occasions as grounded in given cultures and, possibly, group identities[3]. A few scholars moved close to such an approach, in particular Mascarenhas and colleagues (e.g. 2009, 2013a, 2013b; cf. also Rehm, 2010), leveraging sociological literature on rituals. However, they maintained that "activities can be separated into two classes: ritual activities and technical activities. Whilst a ritual activity is described as expressive, rule-governed, routinized, symbolic, or non-instrumental, a technical activity is described as pragmatic, spontaneous, and instrumentally effective" (Mascarenhas et al., 2009: 307). On the contrary, we contend that the expressive order of interaction –with its rituals such as mutual greeting (the example taken in Mascarenhas et al, 2009) or apologies followed by minimizations (cf. further)– is attended to (also) co-occurrently with instrumental activities. In other words, they are not two classes but two dimensions of human activity. This also means that the ordinary interaction rituals we tackle in this article, are as much "spontaneous" as any other situated instrumental activity: they may emerge as required in the midst of "technical activities" (which, in turn, can be as much rule-based as rituals), and should be enacted accordingly to the developing circumstances of that specific scenario and cultural context.

---

[1] Goffman distinguishes the "social occasion" (the scenario, the *kind* of situation) and the "social situation" (a particular, specific, situated instance of a given scenario/occasion).

On the one hand, this approach allows to keep culture into the picture as a multilayered phenomenon, and to avoid stereotyping, whose threat has been largely acknowledged (e.g. Buolamwini & Gebru, 2018; Wang and Kosinski, 2018; Katz, 2020). On the other hand, it pushes further what we mean by socially-competent artificial agent. The perspective we take departs from an individual trait approach to cultural identity, for three reasons. First, because of intersectionality (e.g. Crenshaw, 1991; hooks, 2014), which requires to look at identity as a complex trait function, so to speak, which varies with varying situations. This discloses individually-based cultural adaptation both as nearly impossible in terms of design, and as unethical and dangerous when considering the opportunities and outcomes of developing this kind of technology. Second, several studies highlighted identity's fluid, processal and situated character. One's identity changes both in time (life course) and with changing occasions; in fact, some kinds of "trait" are in the background (e.g. salary, gender) and others in the foreground (e.g. friendship, job expertise) in any given occasion, and this can even change during the event —some of the disruptions we examine in this paper are a case in point. Finally, culture intended as a (more o less) stable property of the individual brought to contradictory results in HRI research (Lim et al., 2021) and it has been noted that

> it is essential to investigate under what conditions could critically benefit from culturally sensitive robots. Studying across cultural practices and learning from broader literature on social anthropology, sociology, ethnography or human–computer interaction would provide exemplars to identify and point to priority areas for future research on social robotics and robotic design. (ivi: 1328).

We maintain that a priority area is constituted by interaction rituals, with particular attention to interactional disruptions.

There is, furthermore, a third kind of context alongside social occasion as located in a given culture, the situated context (see footnote n. 1). Besides bringing in its specific, local developing circumstances, such a situated context also entails the possibility for troubles and disruptions to emerge –at the functional, conversational, and expressive levels. Agents in interaction do not only commit practical mistakes such as dropping the sugar on the bar counter, but also conversational errors and other "infelicities" in communication (e.g., misunderstandings) that equally ask for a "repair" (Schegloff et al., 1977; Raymond et al., 2013; Drew et al., 2015). In both cases, what we are referring to as the functional order of action-in-interaction can be disrupted; in both cases, the expressive order is as much at stake (e.g., appearing clumsy, or stupid). Whereas the former have been addressed in HRI (e.g., Marge, Rudnicky, 2019; Benner et al., 2021), devised solutions, to the best of our knowledge, do not tackle the expressive order, thereby missing to acknowledge one of the dimensions building up to the *social* competence of an (artificial) agent. Moreover, disruptions can directly affect the expressive layer of interaction, in the absence of any material or (intrinsically) conversational trouble, for instance when an "irrelevant" topic penetrates the "membrane" and enters the conversation. This equally calls for a mending, or recovery strategy.

---

[2] E.g., an individual born and/or living (that is rarely specified in HRI literature) in a given country, and from a given generation (Cortellessa et al., 2008; Torta et al., 2014).

[3] "Possibly", as social groups and social gatherings are different entities. Whenever interaction occurs, we have a social gathering, but it is not a given that such gathering involves members of an actual social group with its distinctive culture (e.g. the bar counter at an international airport).

When considering the design and development of *socially-competent, culturally-adaptive artificial agents*, all the above mentioned layers must be taken into account –from a perspective looking at culture as an interactional and fully social, rather than individual dimension–, and artificial agents should be enabled to contribute to recovery strategies concerning "interactional breaches and wider relational ruptures" (Tavory & Fine, 2020: 365). In this paper we argue that the capability of an artificial agent to detect and recover interactional disruptions can be important to equip the agent with effective social skills (skills for the expressive level of social interaction), and that such specific skills can be culturally-adaptive via culturally-located training data-sets for the expressive order of social action-in-interaction. This claim is illustrated by means of a simple applicative scenario (the artificial bartender, see Section 2) that helps us to present a review of the relevant literature on interactional disruptions and their corresponding recovery strategies (Section 3). After having organized the information about the main disruptions in an illustrative table (Section 4), we show (Section 5) as a proof of concept how first a reactive agent architecture, and then an architecture with planning capabilities can detect, recover, and even prevent some disruptions. This will help us to sketch possibile requirements for an artificial agent with respect to the whole list of disruptions, providing suggestions for implementation. We briefly discuss the proposed approach (Section 6) before concluding.

## 2.   Guiding scenario: the (artificial) bartender

The use of embodied artificial agents outside the constrained environment of production sites (industrial robots and cobot) or the controlled areas of medical needs, has focused on helping people solve everyday activities, from interactive wheelchairs to automatic cleaners, from chat bots to automated travel agents. One commonality of these artificial agent systems is the focus on functional tasks, or instrumental behavior. By functional tasks we mean roughly what is captured by the following description: the person has some physical need or a goal to realize, and the artificial agent ensures that the need is satisfied and the goal achieved. For instance, Marge and Rudnicky (2019) looked at navigation tasks and how to solve and improve solutions for miscommunication in task-oriented spoken dialogue. They focus in particular on the indexical nature of human verbal communication in situated interaction ("situated grounding problem").

Consider another example, a bartending scenario, which has been addressed by several AI scholars (e.g. Foster et al., 2012; Petrick and Foster, 2013; Giuliani et al., 2013). If we aim to build an artificial bartender, the focus would be to have a robot that can satisfy the typical requests of clients, like to get some coffee within a certain range of variety. The typical interaction that the agent is designed to carry on would be like the following:

Client: Good morning! May I have a coffee?
AI: Yes, of course. Would you like to have some milk with it?
Client: I'll take plain coffee, thanks.
AI: Anything else for you?
Client: I'd like a glass of water, please.
AI: Very well. That's 3.30 Euro in total.

This interaction can be modeled formally in standard logic formalism and, thus, allows us to build artificial agents that can not only perform but also analyze and learn from these interactions. For instance, the previous ordering scenario is captured by the following set of logical formulas (**c** stands for client; **x**,**x'**,**y**,**z**,**w** for objects; **a** for the artificial agent, the other components of the language are standard):

| | |
|---|---:|
| Client: Request(c, x, a) | Client requests coffee to AI |
| AI: Accept(a, do(make(a, x))) | AI accepts coffee request from Client |
| AI: Request(a, x', c) | AI asks if milk is requested by Client |
| Client: Request(c, x, a) | Client requests plain coffee to AI |
| AI: Accept(a, do(make(a, x))) | AI accepts plain coffee request from Client |
| AI: Request(a, y?, c) | AI asks if something else is requested by Client |
| Client: Request(c, z, a) | Client requests glass of water to AI |
| AI: Accept(a, do(make(a, z))) | AI accepts glass of water request from Client |
| AI: Request (a, do(pay(c, w, a))) | AI requests the payment to Client |

The capacity of artificial agents to plan and perform in social situations like the above has helped to increase awareness of the possibilities of AI in these contexts but has so far not led to consider the actual variety of real scenarios. While the ability of today's artificial agents to correctly participate in interactions is remarkable, the type of scenario in which these agents specialize is deprived of a dimension of social interaction that characterizes people's living together in everyday life. The problem we are raising, in other words, is the limitation of any solution whose focus is limited to solving functional tasks, or to put it differently, limited to task-oriented and often exclusively dyadic interaction. Giuliani and colleagues (2013) did consider multi-party interaction and moved beyond task-based interaction, towards what they call "socially intelligent bartender", yet the expressive order of social interaction remained excluded.

On the one hand, people tend to attribute mood and opinions to agents, artificial or not, with whom they interact, and thus tend to initiate in parallel sociable (i.e. non-functional) interactions with them. Even when this is not so, in everyday life people naturally intertwin functional tasks with purely social –or socially expressive– acts. For example, if we look at what actually happens in a real bar scenario, we find something like the following:

1    Client: Good morning! May I have a coffee?
2    AI: Yes, of course. Would you like to have some milk with it?

| | |
|---|---|
| 3 | Client: Well, I'm tempted since it's already the third today but if you make a really good one I'll take it plain. |
| 4 | (A socially-competent reply could be: You have to taste mine! I'll make it plain then, ok?) |
| 5 | AI: Anything else for you? |
| 6 | Client: Yes, I'd like a glass of water also. It's so hot outside. |
| 7 | (A socially-competent reply could be: They say it will stay above 30C for another couple of days) |
| 8 | Client: I see. By the way, the children outside are very noisy! |
| 9 | AI: Really? From here you cannot hear anything when the front door is closed. |
| 10 | [Client drops the box of sugar on the counter.] Client: I'm sorry. I'm clumsy today, don't know why. |
| 11 | (A socially-competent reply could be: Don't worry. There are days like that.) |
| 12 | AI: That's 3.30 Euro in total. |

This case is much more complex than the previous one and shows the flexibility of social interactions. It makes evident the presence of an "expressive order" (Goffman, 1967) running parallel to the functional one. Focusing on the final lines we can see how the AI contributes to the maintenance of the expressive order by replying with minimizations (line 11) to the client's apologies ("I'm sorry") and excuses ("I'm clumsy…") following a mistake (line 10). Issues such as this are quite frequent during an interaction. In these cases, the agent should intervene to mend the social situation.

Consider another example:

| | |
|---|---|
| 1 | Client: Good Morning! May I have a coffee? |
| 2 | AI: Yes, of course. Would you like to have some milk with it? |
| 3 | Client: No, thanks… |
| 4 | AI: Do you want... [interrupted] |
| 5 | Client: Oh no, I want milk… Sorry, today I'm lost in thought. |
| 6 | AI: Don't worry, it's not a problem. I guess this is a busy day for you. |

The client offers apologies and justifies him/herself (line 5); the AI must accept the apology and contribute to mending/recovering the social situation, shifting the focus to another topic (line 6), which is an example of minimization. That is, to behave socially the AI must take care of the expressive dimension of the interaction alongside its functional one.

Today's artificial agents do not have the capabilities to interact at this level. This is not only a technical issue. We lack a(n interdisciplinary) theoretical framework to make this kind of interaction accessible to artificial systems. The focus is not to enhance efficiency in task-oriented HRI, but to pave the wave for adding this layer of social competence. The rest of the paper will concentrate on the conceptual elements needed to frame these social interactions and aims to show that it is possible to build advanced artificial agents that can be called social agents in this stronger sense.

## 3. The expressive order of interaction

To provide the necessary theoretical and analytical tools, this section draws upon a body of literature in microsociology and more specifically, Goffman's (e.g., 1959, 1961, 1974) interactionist approach, and Garfinkel's (e.g., 1967, 2002) Ethnomethodology (EM), out of which Conversation Analysis (CA) (e.g., Sacks 1992; Goodwin and Heritage, 1990) has emerged. This area of study is currently known as the EM/CA approach. By applying the EM/CA framework within a human-robot interaction perspective, we will be able to provide a categorization of disruptions in social interaction that will be used later to analyze the requirements needed to build socially-competent and culturally-adaptive agents.

Microsociology considers *social interaction as an ordered activity* (the ordered basis of the larger social order). Garfinkel (1967) maintains that human actions-in-interaction are designed to result immediately intelligible and reportable —accountable— to any other co-present member of (a given) society. It is not that we accompany our actions with explicit accounts, that happens only when troubles emerge; we design action to be self-explanatory to our fellow social members based on the context at hand. In discussing the "interaction order", Goffman (1983: 3) talks of people's "capacity to indicate their own courses of physical action and to rapidly convey reactions to such indications from others" and sees this as the precondition for action coordination: "Once individuals –for whatever reason– come into one another's immediate presence, [...] the line of our visual regard, the intensity of our involvement, and the shape of our initial actions, allow others to glean our immediate intent and purpose" (ibidem). He further notices how "[s]peech immensely increases the efficiency of such coordination, being especially critical when something does not go as indicated and expected" (ibidem). Indeed, as ethnomethodologists underlined, the orderliness of social interaction is a collaborative, situated and processual accomplishment by social actors. Therefore, disruptions are always possible, and are to be managed depending on the situation and the relationships holding among participants:

> [P]eople actively strive to construct shared lines of action in order to preserve social relations and their own identities as competent actors [...] But the crucial point is that people must constantly re-calibrate which lines are not to be crossed. Some disruption is a part of social life, but only insofar as it doesn't threaten the underlying fabric of order that people struggle to maintain. (Tavory & Fine, 2020: 368)

Can artificial agents contribute to the interaction order and the management of its disruptions as expected by socially-competent human agents in social situations? Not today. The problem is complex but the main reason, we believe, is that we do not really know if and how an artificial agent can manage this kind of information. Our goal is to show that this kind of knowledge can be classified, formalized and made available for further processing.

Speaking of action coordination is talking of joint action in task-oriented interaction as much as, and simultaneously, talking of *alignment* in the dramaturgical structure of interaction —a shared understanding, a "*common definition of the situation*" at hand (Thomas, 1923; cf also, e.g., Goffman, 1959, 1967; Tavory & Fine, 2020). The expressive dimension of the "interaction order" —or *the expressive order*— has to do with the social occasion, the corresponding social roles, "impression-management", "face work" and connected interaction rituals, such as those concerned with deference and demeanor (Goffman, 1959, 1967; cf. also Collins, 2004).[4] The expressive order is always, although differently, operative no matter the ongoing activity and the practical purposes at hand. It is both a means for coordination and a result of coordinated action. It allows certain degrees and kinds of disruption depending on the social occasion (i.e., the scenario), the situated circumstances, and the relationships holding among participants. However, not all disruptions are allowed —not those that "threaten the underlying fabric of order that people struggle to maintain" (Tavory & Fine, 2020: 368), not those impeding social trust (Garfinkel 1963; Turowetz & Rawls, 2021). Therefore, people cooperate to the maintenance, and occasional repair/recovery, of the expressive order of interaction when it gets threatened "beyond limits". Moreover, "[w]hen instrumental goals are clear, as are interactional scripts" —the bartending case, is an example—, "disruptions are relatively rare [...] In these *casual circumstances*, [...] interactants worry that their interaction is so fragile that it will dissolve at any sign of disruption" (ivi: 378, emphasis added). The socially-competent artificial agent has first to understand that there are two dimensions, superimposed one on the other and managed simultaneously in social interaction —functional and expressive— and then learn about the latter too.

---

[4] Also notice that the expressive order dynamics significantly affect the emotional and affective dimension of interaction, an aspect developed in particular by Collins (2004).

There is "a profound cultural know-how that actors bring into interaction" (ivi: 381). Culture is *orthogonal* to both dimensions. The way things are done in a situational scenario —i.e., the ethnomethods (Garfinkel, 1967)— culturally varies. However, there are reasons to believe that cultural differences affect the expressive order of social interaction more significantly than its functional dimension. For example, the degree of deference and the ways in which deference is expressed —i.e., displayed in interaction— greatly vary among cultures and contexts. Take for instance clerk-customer relation: *deference rituals*, for each role, vary with contexts (e.g., the friendly, easygoing clerk in New York, opening conversation with a "How're you doing"; the embodiedly deferent clerk in Tokyo) as well as other culturally-marked characteristics of the context, such as status and prestige (e.g., a clerk of an Armani store in Manhattan maintains a similar degree of non-intrusiveness of the clerk working in an average-shop in Tokyo, although likely, the former would not embody deference as much as the latter). To act in a culturally proper way, therefore, a social artificial agent must learn how the expressive order is managed in that context.

A peculiar element of the expressive order are the *rules of (ir)relevance*. According to Goffman (1961), social encounters are characterized by an enveloping membrane which, following the cell analogy, keeps outside "issues", or contents, topics considered irrelevant for the current encounter. Filtered out "issues" are, chiefly, social categories (es. status, class, gender). The ways in which social members are categorized in interaction (cf. Watson, 1978; Fitzgerald & Housley, 2015) changes with the kind of encounter (i.e., social occasion, or scenario) and with cultures, here intended as properties of situations and groups, rather than individuals. For instance, as we mentioned, in a party at a friend's place in Western societies, social class is usually filtered out; otherwise, the encounter does not properly qualify as that kind of encounter —that is to say, otherwise the interaction order is broken, or risks so, and has to be repaired, or the "frame" of the encounter (Goffman, 1974) will change (indeed, humans can do "footing" (Goffman, 1981) between frames). Regarded-as-irrelevant issues have no citizenship as explicit arguments of talk; it is not that status and prestige do not play any role, but when considered as individual characteristics, they do not feature as explicit, legitimate elements of interaction in given contexts in given cultures. The membrane is actively put in place every time by participants, although tacitly. It is an interactional work in which we are almost constantly engaged when we are with others. This is also the place for allowed disruptions, i.e., "disruptions-for" rather than "disruptions-of" (Tavory & Fine, 2020) the social interaction and connected relations. However, "disruptions-for" are much more frequent in social encounters entailing close relationships than on casual interactional occasions.

Another dimension of the expressive order appearing as particularly relevant for the current discussion has to do with *(dis)preference* as a social (rather than individual or psychological) element. Disagreement is usually dispreferred (Pomerantz, 1984) as a possible offense to the other's face, and as such it is systematically remedied, that is, prefaced with: silence/pause in talk (hesitation), repair initiators such as "hmm", requests for clarification ("Eh?", "What?") or for repetition ("I'm not sure I got it"). There are other examples of dispreferred social actions; for instance, agreement itself is dispreferred when following a self-critique. Such actions always risk breaching the expressive order, that is why they are remedied beforehand when possible. When not possible, a recovery strategy is employed afterwards. What is offensive in one culture could be not offensive in another. For instance, in Anglo-American speaking cultures, talking simultaneously is offensive (individual turn taking system), the same does not hold in Italy on many occasions (cf. Bassetti & Liberman, 2021).

Non-receptions and misunderstandings are other possibly disruptive elements in talk-in-interaction and are called *repairables* by conversation analysts. Misunderstandings are generally managed as follows: the recipient(s), unaware of the problem, offers a reply which proposes an (inappropriate) interpretation of the previous speaker's words; making the problem explicit is this previous speaker's responsibility, and that must be carried out with utmost carefulness with respect to the recipient's face. In the following example, formulation B (line 3) by the artificial bartender is much better from this perspective than the other one.

1    AI: Would you like to have white or brown sugar?
2    Client: Yes, please.
3    AI: A) Both of them?
        B) Do you prefer brown sugar then?
4    Client: Yes thanks. [Or: No, white sugar please. Sorry.]

Also notice how formulation B allows the client not only to "overrun" the issue, but also to self-correct, and/or to apologize for the lack of clarity. In short, opportunities for saving each other's *faces* –an endeavor we are always engaged in when interacting– are enlarged. Apologies and excuses are preferred ways to bring about such an endeavor (Brown & Levinson, 1978; Horodeck, 1981; see also Miller et al., 2009 for an application in conversational agents). In principle, we may also want an artificial agent being able to save its "face", for instance when dealing with ambiguity and situated grounding problems during navigation tasks (cf. Marge and Rudnicky, 2019).

## 4. Disruptions and their recovery

The aim of this section is to collect in a single table the kinds of disruption, with their associated recovery strategies, which may occur in a human-human interaction to make them available, as a sort of organized chart, for detection and subsequent repair. After all, if the artificial agent wants to be validated as a social agent, it must understand the disruptions and know how to repair them in a given cultural context. Therefore, considering an artificial agent acting in a social context, the agent's knowledge module may use this new kind of information to interpret the ongoing social interaction, possibly to identify cases of disruption, and to activate suitable recovery actions which are socially acceptable or, depending on the situation, even expected.

Table 1 is divided in two sections: the first is about functional disruptions, the second about social ones. Each section lists several types of disruption. The columns analyze various aspects of the disruptions. First, a disruption is presented in general terms (e.g., an agent interrupts the functional procedure). A specific example is given at column 5 (e.g., the client does not pay his/her bill). The other columns explain the reasons why this action is a disruption, whether it is intended or unintended (and how it is generally perceived by other people), the agents involved, the functional or social status of the disruption, i.e., its degree of acceptability and the need vs. opportunity for a recovery strategy. The last column reports possible recovery strategies.

The functional part of the table is elicited from the guiding example of the bartender that we employ in this article. However, the considered kinds of disruption (e.g., performative mistake, row 5 in Table 1; lack of response from a human agent, row 3) can be applied to other scenarios. Even though the functional interaction is analyzed in general terms, we do not claim it covers all possible cases as further functional dependencies may occur within other tasks. The second part looks at social disruptions and interactional aspects, or "troubles", concerning the expressive order (perhaps with consequences on the functional order). Note that the part on functional disruptions is more easily exploitable in today's architectures for artificial agents since these have the capability to control the execution of a procedure within the usual sense-perceive-plan-act loop. The part concerning social disruptions presents similar information but, due to the generality —ubiquity one may say— of this kind of disruption, it may look that its exploitation in real scenarios is less clear. However, both in the table and what follows we provide examples of instantiation of the kinds of social disruption we consider.

**TABLE 1. Functional and social disruptions, and their recovery strategies**

| | FUNCTIONAL DISRUPTION | REASON (why this action is a functional disruption) | INTENTIONALITY AND INTERPRETATION:<br>- Intended/unintended<br>- Perceived as intended/unintended | DISRUPTION FUNCTIONAL STATUS:<br>- Processually acceptable<br>- Recovery is necessary<br>- Recovery is optional | DISRUPTION IN THE GUIDING SCENARIO (description of possible happening) | WHO/WHAT IS INVOLVED IN THE DISRUPTION (beyond the artificial agent/s) | DISRUPTION RECOVERY STRATEGY |
|---|---|---|---|---|---|---|---|
| 1 | An agent asks to change what agreed upon in a previous interaction (e.g., change of order, change of payment method, change of type of purchase at the counter/table/to go). | The linearity of the functional process is lost. The process moves back to a phase already concluded with possible difficulties such as material waste and time loss. | The request to change is intended. The disruption is perceived as unintended (the agent made a mistake or changed her mind. S/he did not place the wrong order on purpose). | The status of the disruption depends on the consequences. E.g.: an agent makes an order and then asks for a variation which is compatible. The client orders a latte, then changes and orders an espresso (not compatible). An agent makes an order and then asks for a different thing which requires a different activity (not tolerated): a recovery action is necessary. | The client orders an espresso and then changes to a caffè macchiato (compatible). The client orders a latte, then changes and orders an espresso (not compatible). | Co-present human agents, involved objects (latte, espresso). | 1) The client gives explanation and/or excuse ("I don't know what I was thinking");<br>2) The bartender or client minimizes ("Well, don't tell me this never happens");<br>3) The client proposes compensation ("I'll pay for the latte also"). |
| 2 | An agent doesn't start the functional process setting but no functional process is executed (e.g., in a bar, s/he goes to the counter but does not order anything). | There is a functional setting but no functional process is executed (nor selected) because the input is missing. | The disruption can be involuntary (the agent is absent-minded) or voluntary (e.g. the agent, perhaps assuming the other is busy, expects to be asked what s/he wants to order or is unsure about what to order). | A short delay is tolerated. A recovery action is optional (functional place) but does not make an order. A recovery action is necessary if the functional process is not initiated at all. | The client enters the bar and goes to the counter but does not make an order. | Co-present human agents. | 1) The client justifies the delay in ordering ("Sorry, I was distracted");<br>2) The bartender initiates the process ("What would you like?");<br>3) The client gives explanations for not starting the process ("I just realized I'm late. I must go, sorry."). |

| | | | | |
|---|---|---|---|---|
| 3 | An agent does not proceed with the functional process (e.g., s/he does not answer a question needed to move forward in the process, a possible consequence of non-reception, see row 10). | The missing answer interrupts the functional process. The disruption can be involuntary (the agent is absent-minded, distracted or has not heard) or voluntary (e.g., the agent finds the question or attitude not appropriate and ignores the request). | A recovery action is necessary for the functional process to proceed. A recovery action (excuse, explanation) is optional if the agent decides to abandon the process (see next row). | The client does not answer the question asked by the bartender (e.g., about milk or payment). | Co-present human agents. | 1) The bartender repeats the question; 2) The client justifies the lack of reaction (e.g., non-reception or misunderstanding: "Sorry, I can't hear you. It's noisy in here"; "Sorry, I misunderstood"); 3) The bartender or client minimizes. |
| 4 | An agent abandons the functional process (e.g., leaving before the process completion, possibly stating that s/he has changed mind). | A functional step of the process is not executed with consequent disruption of the process. | The disruption could be intended (the agent does not want to do the next action) or unintended (the agent forgets to do the next action). The missing action can be perceived as intended or unintended by the other agent. | The outcome is not processually acceptable. An act of recovery is required to complete the functional process. | The client leaves before paying the bill. | Co-present human agents. | 1) The bartender asks to proceed with the process ("Have you already paid?"); 2) The client justifies the missing action ("I was distracted", "I thought I already paid"); 3) The bartender minimizes ("Today coffee is free", with a smile). |
| 5 | Functional incompetence, performative mistake. | The situation changes and a new goal (e.g., cleaning after something has been spilled) gets priority, the ongoing process is on hold. | Unintended. | The outcome is processually tolerated. An act of recovery is required to complete the functional process. | The bartender spills sugar or coffee. | Co-present human agents, mismanaged objects (sugar/coffee). | 1) The bartender apologizes for the disruption ("I should be more careful"); 2) The client minimizes ("No problem, it happens sometimes"); 3) The bartender acts to restore a suitable situation (e.g., sweeping the sugar up). |

| SOCIAL DISRUPTION (disruption of the "interaction order" and its expressive dimension) | REASON (why this action is a social disruption) | INTENTIONALITY AND INTERPRETATION:<br>- Intended/unintended<br>- Perceived as intended/unintended | DISRUPTION SOCIAL STATUS:<br>- Socially acceptable<br>- Recovery is necessary<br>- Recovery is optional | DISRUPTION IN THE GUIDING SCENARIO (description of possible happening) | WHO/WHAT IS INVOLVED IN THE DISRUPTION (participants) | DISRUPTION RECOVERY STRATEGY |
|---|---|---|---|---|---|---|
| 6 Disruption of the "*common definition of the situation*" (Thomas 1923), e.g., somebody behaves as implying an intimate relationship between two strangers; someone approaches a passer-by as if they were close friends; somebody asks for an item of the wrong kind (e.g., to ask for socks in a bar). | The action disrupts the *social interaction founding principle* (cf. Thomas 1923). | The disruption could be perceived as unintended, as emerging out of the behavior of an incompetent social member (e.g., child, "mentally ill" person). Alternatively, it could be perceived as intended (absence of social trust, cf. Garfinkel 1963; Turowetz & Rawls, 2021). | If the disruption is *understood as intended*, a recovery strategy is necessary.<br><br>If the disruption is *understood as unintended* (due to incompetence), a recovery act is optional. It is likely to occur if there is a competent third social member.<br><br>N.B. In this kind of cases, where the "social status" of a disruption is ambiguous, the artificial agent should see if the other present humans interpret the disruption as intended or not, and it should then act accordingly. | The bartender, referring to two clients who are strangers to one another, says "Here two lattes for two lovebirds".<br><br>An agent says to another agent who is a stranger to him: "Hey! How are you doing? Has your wife fixed that problem? By the way, I've managed that issue with the dentist." | Co-present human agents. | If the disruption is *understood as intended*, a recovery strategy in casual interactional occasions consists in acting like the disruptive action and its agent are irrelevant (e.g., the two mutually stranger clients laugh together at the bartender or shake together their heads). In interactional occasions entailing closer social relationships, recovery can be needed or not, as a "disruption-for" (Tavory & Fine 2020) could be in place.<br><br>If the disruption is *understood as unintended*, the interactional "move" consisting in acting as if the fact is irrelevant, takes the form of "triangling" (Cain 1983) and/or relies on a secondary semiotic system in the ongoing conversation (e.g., gestural if the conversation is mainly verbal). |

| 7 | Disruption of tacit norms and related expectations (i.e., *ethnomethods* - cf. Garfinkel, 1967): e.g., 1) way of queuing; 2) tone of voice. They depend on culture and context. | Ethnomethods (usual ways of doing things in a given scenario) make other people's actions accountable and involve expectations that help humans to understand the surrounding reality. Behaviors misaligned with expectations are not completely accountable: the action is not as clear as the expected one would be to the eyes of any bystander - a matter of understanding/grasping what's happening (e.g., that an agent is going to the toilet). Similarly, it could be perceived as intended or unintended. Such *lack of accountability* decreases trust among social members (cf. Garfinkel, 1963). | The act could be intended or unintended. 1) The act is intended. The client believes that the motivations for which s/he is not following the social rules are correct; or s/he thinks/hopes the other people won't notice. 2) The act could be intended (there is a lot of noise and the agents may have problems hearing each other) or unintended (the agents do not realize they have raised their voice). | The action ranges from socially tolerable (e.g., tone of voice up to a certain level) to requiring a recovery (e.g., skipping a queue, adding "I'm in a hurry". People expect the locally responsible person (e.g the place owner or keeper), if any, to intervene to restore social order. Alternatively, they intervene themselves (rare, if a responsible person is present – however, should a "responsible" robot be present, cf. also Pitsch, 2016). | A client passes another client at the counter queue, and s/he orders adding "I'm in a hurry". Two clients chat loudly —too loudly for the context— at the counter. | Co-present human agents, engaged in common-and-expected ways of behaving (ethnomethods) unfocused interaction and/or common-focused or jointly-focused interaction (cf. Goffman, 1961; Goffman 1966, Kendon, 1988). | An agent makes explicit the common-and-expected ways of behaving (ethnomethods) — e.g., the bartender says "I understand but even those who were there before you are in a hurry". Ideally, after this kind of action, the client offers apologies and the other agents offer minimizations about what happened. |

| 8 | Disruption of *proxemic norms*: e.g., 1) breaking offensive to the others' of the "withs" (Goffman 1983); 2) excessive proximity to another agent with whom there is no previous relationship. (N.B. The determination of proxemic norms such as "excessive proximity" depends on culture and context.) | The act is perceived as intended or unintended selves. In 1) the displayed relationship does not get acknowledged; in 2) the "individual sphere" around one's body is violated. | The act can be perceived as intended or unintended depending on the circumstances. *Intended*: an agent thinks that the circumstances allow an exception to the proxemic rules which are usually required in that context (e.g., crowded bar, the bartender must serve a table s/he could reach only by passing through two interactants or passing very close to someone). *Unintended*: an agent finds her/himself in the middle of a small interactional group – which s/he did not notice before – or s/he trips and bumps into someone. | A recovery strategy is necessary. | Co-present human agents engaged in unfocused interaction, elements of the environment. | 1) A client, or the bartender who is waiting on other clients, gets in the way of two people who are visibly together (e.g., they are talking to each other or engaged in any other "display" of being together). 2) A client reaches too close to or touches another client at the counter or at the cashiers. | Apologies and/or excuses (cf. e.g., Brown & Levinson 1978; Horodeck 1981) by the client who breaks proxemics norms, ideally followed by minimizations offered by the other agent/s. Examples: - *preventive or simultaneous*: A1 "I am sorry, I have to pass through" - A2 "Of course, yes"; - *subsequent*: A1: "I beg your pardon, I didn't see you. Today I am absent-minded" - A2 "Don't worry. It happens to everyone". |

| 9 | Disruption of *conversational norms*: e.g., 1) the second part of an "adjacent pair" (question-answer, request-acceptance, etc.) is missing; 2) one or more of Grice's conversational "maxims" are breached. | A disruption happens because tacit norms relating to conversation are breached, with possible consequences on the functional, alongside the expressive order. | The act can be intended or unintended (e.g., the chatty agent does not notice that other agents stand in a queue because that is behind him/her). However, the act is perceived and socially *managed as intended* because talking is an action socially considered as intentional (so even if it is unintended, we treat it as intended because the alternative would be a worst offense to the agent's "face"). | The act is socially acceptable (up to a point). A recovery strategy is optional; if employed, it can ask for (may-form) or require (must-form) a reciprocating act – see next column. | 1) A client at the counter asks for a coffee and the bartender goes away without answering (then s/he serves the coffee). 2) A client at the counter drones on about personal anecdotes/stories, thereby holding the bartender longer than expected (effects on the functional order). | Co-present human agents in conversation with each other (i.e., engaged in jointly-focused interaction). | Apologies and/or excuses ex-*post* (subsequent) by the agent who breaches conversational norms, and who might suddenly realize that is the case. Ideally (may-form), this is followed by minimizations on the part of the other agent/s. Recovery enacted by the "recipient/s" of the disruptive action (e.g., Bartender: "Excuse me but there are too many people waiting for me to order."). In this case, a reciprocating act is needed (must-form) from the "breaching agent" (e.g., "I'm so sorry. This is a bad day, pardon me."). |

| | | | |
|---|---|---|---|
| Conversational "repairables" (e.g., Raymond et al. 2013): 1) Non-reception: an agent does not perceive what the other agent said; 2) Misunderstanding: an agent misunderstands what the other agent said (the latter is the only one aware of this); 3) Speech error: an agent makes a mistake in talking (uses a word or expression not corresponding to what s/he wants to communicate, or does a phonatory mistake in voicing); 4) Indexical, or situated grounding problem: ambiguity of referent potentially bringing to non-understanding, with possible effects on the functional level. | A disruption might happen if conversational "troubles" are not dealt with properly, with possible consequences on the functional, alongside the expressive order. | The disruption is unintended and perceived as such in all four cases. Cases 1) and 4): A recovery strategy is necessary to avoid the impasse (functional and social level). Cases 2) and 3): A recovery act is optional but could be instrumental to the maintenance of the functional and/or the expressive order (e.g., to avoid receiving an item/service which is not the one needed). | Co-present human agents in conversation with each other (i.e., engaged in jointly focused interaction); elements of the environment (surrounding phenomenal field). 1) The bartender does not hear what is said by the ordering client. Or the client does not hear the bartender's question (e.g., "Do you want some milk or sugar?"). 2) The bartender misunderstands the order and therefore asks a "silly" question afterwards (e.g., "Do you want some milk or sugar?" – "Ehm, no, I asked for a Coke"). That's when the client understands that there was a misunderstanding and can clarify. 3) The client says "I'll have a Coke" instead of "I'll have a coffee". 4) The client asks "Can you please fetch me one of those?" and the bartender wonders what (napkins, sugar pots, spoons …all on the bar counter). | The recovery strategy consists in a "repair" which can be accompanied by apologies: 1) Asking to repeat – must-form of clarification, by the agent who does not receive the message; 2) Signaling a misunderstanding – may-form of clarification by the agent whose words were misunderstood; 3) Self-repair by the agent who committed the speech error; 4) Asking for clarification, either explicitly or implicitly, by the agent who does not understand the message |

| # | Description | Condition | Context | Example | Strategy |
|---|---|---|---|---|---|
| 11 | Disruption of the "membrane" (Goffman 1966), which are the "rules of (ir)relevance" related to the topic of conversation: an agent introduces a "sensitive" topic for that kind of social encounter (e.g., one participant's salary). | There is a social disruption because some topics should remain tacit according to understanding of the social occasion (scenario) and the social situation (situated context). Alternatively, the action could be *intended* (e.g., to embarrass the other agents or make trouble, or in case of "disruptions-for"). Similarly, the disruption could be perceived as unintended and caused by inattention (and can be recovered without pointing it out) or the disruption could be perceived as intended (and it requires to be managed). | The act is not socially acceptable and a recovery strategy is necessary (otherwise the "frame" (Goffman 1974) changes, that specific context, e.g., status, class, gender, religion, sexual and/or focused interaction, and/or bystanders (i.e., unfocused interaction). | A client mentions an aspect, or dimension of social life which is usually "filtered out" in each other (i.e., engaged in jointly-focused interaction), goes on talking/acting as nothing happened; - Ironically commenting or joking in a benevolent way about what the speaker said: again "filtered out"; - Criticizing the disruption (mockingly or dryly, not in a benevolent way). | - Pretending/ignoring the disruption happened (people act as if the topic was not mentioned and everybody goes on talking/acting as nothing happened; - Ironically commenting or joking in a benevolent way about what the speaker said: "disruptions-for" recognized as such, it is employed also in casual interaction as it allows to focus on the disruptive action rather than on the "sensitive" topic, which is again "filtered out"; - Criticizing the disruption (mockingly or dryly, not in a benevolent way). |
| 12 | Carry out socially *dispreferred actions*, e.g., explicit expression of disagreement; explicit expression of disagreement—i.e., verbalization—of disagreement (or missed agreement) following self-criticism expressed by another agent; refusal (vs acceptance) of a request. | There is a disruption due to an offense to the other agent's "face" (Goffman 1959, 1967). | The action is mostly intended, and is always perceived and socially managed as intended. As a matter of fact, even when the action is unintended (e.g., the agent is absent-minded and s/he "speaks before s/he thinks"), it is managed as intended because talking is an action socially considered as intentional (so even if it is unintended, we treat it as intended because the alternative would configure as a worst offense to the agent's "face"). | A recovery strategy – called "remedy" – is necessary unless someone aims for open conflict. | Following an agent's claim, another agent shows explicit disagreement. E.g., bartender:"Please, be patient, now there is confusion" - client: "Well, you should be able to manage rush hour!". Following self-criticism from an agent, another one shows explicit agreement. E.g., the client changes his mind: "An espresso, thanks. Indeed, a latte please, I changed my mind. Today I'm fickle ((laugh)) - the bartender replies: "Huh, it isn't | *Ex-post remedy*: minimization (e.g., "but then everyone has his own opinion, no?") from who produced the dispreferred action, preferably followed by minimization by the other agent. *Preventive remedy*: pause in talk and/or hesitation and/or so-called "remedy initiators" such as "hm" with which the agent prefaces his/her dispreferred action (this also provides the other agent with the opportunity to adjust, rephrase, better explain her/his opinion). |

| |
|---|
| |
| |
| |
| |
| good not to know your own desire!"). |
| |
| |

At the most general level, people in interaction need to know in which situation they are, which is the current social occasion or scenario at hand. In short, they need to tacitly share a "common definition of the situation" (Thomas, 1923). With this, they also know which are the local "methods" —i.e., ethnomethods (Garfinkel, 1967)— to bring about action in interaction in the given context (which is also to say, they know how to instantiate that context/scenario). Particular "classes" of ethnomethods, which we further specify in the table, concern, on the one hand, a fundamental layer of copresent social interaction —namely, proxemics— and, on the other hand, talk-in-interaction, that is, conversational norms and expectations as well as conversational "repairables" (e.g., Raymond et al., 2013; Drew et al., 2015).

Non-receptions, speech errors and especially, misunderstandings and ambiguities (repairables) should be handled properly to avoid functional disruptions and to manage the expressive order of interaction. As Benner and colleagues (2021) recently recognized:

For conversation design, appropriate improvement loops must be considered, which allow both the CA [conversational agents] and the user to fix misunderstood statements or ambiguities [...] Further, prior research shows that users are less frustrated if systems such as CAs apologize for errors.

This concerns dealing with "troubles" that have manifested in talk and that are known as possible trouble when in conversation. Known are also the methods for repairing such problems, and the extent to which a repair is necessary: non-receptions require it, as a clarification is needed to proceed with/in the interaction (this is called a must-form of clarification), whereas misunderstandings are managed in more varied manners (may-form of clarification) as much as speech mistakes.

On the other hand, conversational tacit norms may be breached, such as Grice's "maxims"[5] or the expectation concerning the second part of an "adjacent pair" (e.g., question-answer, greeting-greeting, request-acceptance/refusal). In this case, it is not that a problem occurred during a joint activity such as conversation (more precisely, talk-in-interaction) and the parties have to manage that; rather, one of the participants does not behave as expected by a competent social members, and the parties have to decide whether to make that an issue in the current interaction or not, depending on the situated circumstances (a recovery strategy is optional; for instance, it might be employed to avoid functional disruptions, like the bartender saying to an overly chatty customer: "Excuse me but there are too many customers I need to attend to").

Ethnomethods can be also thought of as to include the topic and content of conversation, not only in general terms (which categories of topics are common or usual, hence can be expected, in a given occasion - see row 6 in Table 1), but also and especially in negative terms (which categories of topics should be filtered out - row 11). Such "borders control" is aimed at defending participants' "faces" and avoiding "disruptions-of" (Tavory & Fine, 2020) social relationships; the objective chiefly concerns the expressive order of interaction, rather than the functioning of talk- or more generally action-in-interaction, or the functioning of the instrumental activity at hand (e.g., ordering a coffee at the bar). Conversely, the maintenance of the "membrane" of a social encounter also maintains the encounter as one of a particular kind (e.g., friendly party), thereby sustaining the common definition of the situation which is currently operative in a given interaction.

---

[5] Benner and colleagues (2021), actually consider one of Grice's maxims, the quantitative, but read this as the following: "the statements in interaction with CAs should basically be formulated as briefly and concisely as necessary [...] So, minimizing required dialogue while maximizing its effectiveness, ensures that mis- and non- understandings are minimized as well. In the context of recovery strategies, this implies that these strategies must also minimize the necessary dialogue." This is a misreading of "minimization" as understood in Conversation Analysis, and in our Table 1.

Finally, there are actions (expressed in talk and/or by other means, e.g., gaze) that are dispreferred and risk causing a disruption of social relations. The matter is not how an action is brought about (e.g., if the talk is audible or not, concise or overly extended), but what an action is —the category of action, what we do with/via an action (e.g., via talk). Being (dis)preferred is not an intrinsic property of any kind of action, nor it depends on its un/skillful execution, but on the local developing circumstances: for instance, as we mentioned, agreement is generally preferred but when it follows a self-critique, it is not. When dispreferred actions cannot be avoided (for various reasons, e.g. expressing one's opinion, or "saving" the functional activity at hand —serving at a bar— in front of a chatty customer), several strategies can be employed beforehand to mitigate its impact on social relationships and on the co-participant's "face", to the point of allowing them to adjust their line of action (e.g., modify/clarify a previous statement by rephrasing).

## 5. Agent Architecture

The case study we considered includes a possible embodied artificial agent that interacts with a user in a context in which a functional performance is expected from the first (provision of a good, the coffee, in exchange of money). To this end, the relevant functional part of the artificial agent is the dialogue system, that can engage the user in a dialogue in natural language. In general, the scope of a naturally occurring interaction is not constrained from the side of the user, who is likely to produce sentences partly outside the narrow domain of the task-oriented functional interaction. Following the survey of Deriu et al. (2021) on the evaluation of dialogue systems, we can divide such systems in three main sets: task-oriented, conversational, and Q&A. In all these, we have recently witnessed a shift from complex modular architectures to end-to-end data driven approaches powered by Deep Learning and Deep Neural Networks techniques. An analogous phenomenon is the growing interest in the application of Machine Learning, and in particular the Reinforcement Learning paradigm, to robotics. However, it is reasonable to expect that these competing architectures will coexist for some time in the future and, more generally, will evolve into hybrid systems that can take advantage of the strength of each approach.

We argue that the aspect we are focusing on in this paper –namely the management of the expressive order, over the functional one, during an interaction– can be usefully included in task-oriented dialogue systems without requiring the power and computational resources needed for a pure conversational agent. In other terms, we would like to add to the task-oriented agent just enough social competence to maintain the interaction and lead to its functional goals, exploiting and implementing the mechanism of disruption detection and consequent disruption recovery mentioned above. This mechanism could be exploited also in conversational systems, even though we do not investigate this specific area further. In order to provide a dialogue system with such social functionality, we need to consider separately the case in which the system is realized with an end-to-end data-driven architecture (based on Machine Learning) and the case in which it is instead devised with a modular architecture.

In *End-to-end Data-driven Architecture,* this case the considerations of the importance of the expressive order have a role in the process of the selection of the data with which to train the Machine Learning models. What we foresee is that the conceptualization we propose can be used to tag and select the data in a way that is task-dependent (for disruption detection) and culture-specific (for repair selection) and the data can be included in the training set. Note that the pairs disruption-repair can be included in the normal task-oriented data, given that disruptions may happen anytime, and their effect is controlled and smoothed with the repair mechanism. In this way, the resulting module can also gain this culture-dependent adaptive mechanism. A practical architecture of such a system can, for example, learn the disruption-repair task separately or be completely integrated, with the disadvantage, in the latter case, of having to re-learn the module for each different culture deployment. It would be also possible to consider the application of Reinforcement Learning; however, the definition and the generation of the reward function does not appear to be trivial, because it would involve the direct social feedback from human trainers/users who would be involved in long and possibly frustrating interaction with the agent. Moreover, among humans, this kind of "rewards" concerning the expressive order of interaction are dealt mostly tacitly (e.g., smiling, or simply not raising issues of sorts - cf. e.g., Goffman, 1983).

In a *Modular Architecture*, the agents should be equipped with at least two additional components: i) the disruption detector and ii) the recovery component. One expects the detector to be informed about the nature of the task. It could be seen as an anomaly detector for it. On the other hand, the same recovery component can be used for a range of tasks. Nonetheless, it should be tuned to the culture where an embodied version is deployed. To this end, several solutions seem to be possible: the detector could be realized in a data-driven way, by implementing an anomaly detector for the task, whereas the recovery component can be driven by rules and knowledge or by a library of reparatory actions-in-interaction among which to choose the most appropriate one. The effort of developing the recovery component is that it can be used across-tasks and can be useful to adjust dialogue systems with respect to the culture at hand. Both the modules should be integrated in the modular architecture.

What is described here has the goal to point out how the expressive level can be usefully included in existing dialogue systems architectures, to contribute at smoothing the social interaction within a task-oriented dialogue system. Obviously, what we propose here would need to be implemented and tested to verify on the field the technical realizability and the quality of the user interaction.

To show how the requirements emerging from Table 1 could be satisfied in a realistic modular architecture, let us consider the adaptation needed in the particular case of the architecture presented by Pustejovsky and Krishnaswamy (2021). Dialogue systems are often disembodied, yet our proposal requires considering also physical interaction with the world, and the architecture cited above fulfills this requirement: Pustejovski and Krishnaswamy present an embodied agent that considers the dialogue system and embodied interaction holistically, by modeling interactions by means of communicative acts.

In our case, the connection between the requirements and the architecture is given by: 1) a multicultural library of disruptions and a multicultural library of kinds of action –particularly, communicative acts– that help recovering from disruptions; 2) a culture-specific function of disruption recognition; 3) a culture-specific map between disruptions and recovery acts; and 4) a culture specific module of the execution of the act. Figure 1 depicts the modular architecture for disruption recognition and recovery based on a library of communicative acts (Architecture A). The idea is that two libraries (that can be also thought as ontologies) of, respectively, disruptions and communicative recovery acts are devised. Both libraries are assumed to be multicultural, in the sense that they represent disruptions and communicative acts in different cultures with cultural annotations. The other part is devised for a single culture and comprises three modules, *Disruption Recognition* (DR), *Communicative-Act Execution* (CAE) and a third module called *Map Disruption Communicative-Act* (MDCA) that maps in a culture-dependent way disruptions into communicative acts playing the role of connection between the other two modules.

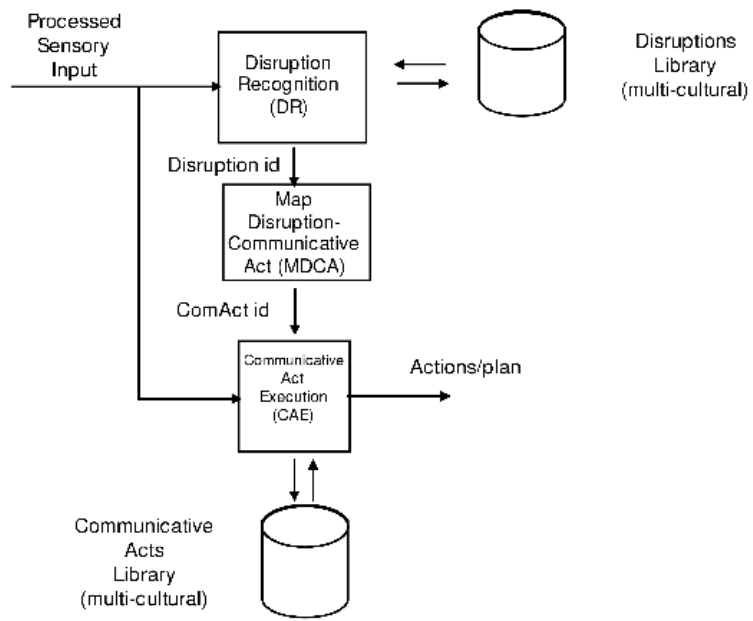

**Figure 1.** Modular Architecture for disruption recovery (Architecture A)

Architecture A provides recovery for disruptions originated either by the artificial agent (failures) or by the other participants. For example, let us consider a case depicted in the first row of Table 1 (where the functional disruption caused by a change with respect to what declared previously in the same interaction is introduced). Here the customer orders a latte by error and then changes her mind. In this case, one possible recovery strategy is *minimization*. The DR module recognizes the change of the order, if it happens within the window of the processed sensory data, and consequently the disruption. The module MDCA maps the disruption to the communicative act of minimization. Finally, the third module CAE produces the actions or plans necessary for the minimization. For example, uttering the sentence "Never mind, it is not a problem".

As a second example of how Architecture A would work, let us assume that the artificial agent spills the coffee on the counter, an event corresponding to row 5 of Table 1, of the functional disruptions part of the table (where incompetence and/or performative mistakes are considered). The DR module recognizes the spilling and consequently the disruption. The module MDCA maps the disruption to the communicative act of *apologies*. Finally the third module CAE produces the actions or plans necessary for the act of apologizing, for example uttering the sentence "I am sorry". In the meanwhile, the physical planning of the agent (not shown in the architecture) should plan and control the remedial physical action(s), namely cleaning and providing another coffee.

Overall, Architecture A can be thought of as reactive; therefore, it does not take into account the undesirable possibility of generating further disruptions. In fact, a simple reactive architecture can effectively respond to some of the disruptions, but the lack of reasoning and representation eventually could prevent the successful handling of many disruptions. A more complex agent architecture must be devised in order to permit it to reason about the potentially disruptive consequences of its (recovery) actions.

Figure 2 shows a more general modular architecture (Architecture B) that prevents disruptions based on the presence of a *Planning Module* (PM). As a step forward with respect to Architecture A, Architecture B includes a *Enhanced Disruption Recognition* (DR) module; moreover, the mapping is now between disruptions and recovery strategies by means of the module *Map Disruption Recovery Strategy* (MDRS), and the actual execution of the actions is entrusted to the *Action Execution* (AE) module. The underlying assumption is that the world representation used by the planning module includes representation of *scenarios* and, in particular, of the *current scenario* as well as the consequences of actions on the current scenario whose status is maintained by components of the agent (these are not shown in the figure). The current scenario is in fact used by the recognition module EDR to detect a broader range of disruptions. The planning module PM is exploited to prevent further disruptions. In this case, the PM aims at repairing the disruption while avoiding generating further disruptions, and plans accordingly. Moreover, it is possible to detect beforehand disruptions that depend on a change of the state foreseen as a consequence of an action.

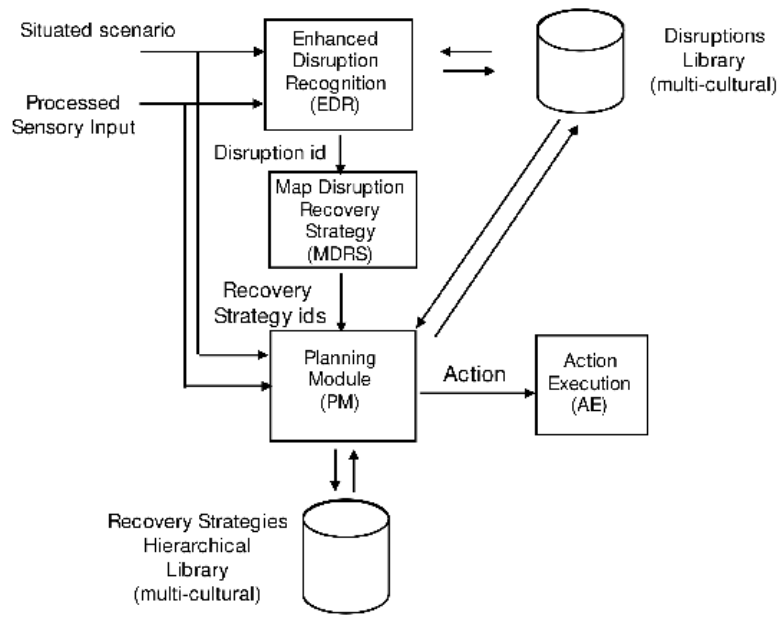

**Figure 2.** General modular architecture for disruption prevention and recovery based on a planning module (Architecture B)

To exemplify the workings of Architecture B, consider the first example presented above for Architecture A, where the customer changes her mind and wants a coffee after placing the order of a latte (row 1 in Table 1). We assume that Architecture B manages the current situation of the client order and the disruption due to the change of order. In this case, the EDR module keeps track of the developing situation and places a second order. The disruption is mapped by the MDRS module to a recovery strategy, namely *minimization*, that is inserted as a goal in the PM that plans the appropriate action (saying "Never mind, no problem at all") and sends it to the action execution module. The assumption that the change of the order placed by the customer has to be done within the window of the processed sensory input, done with the simpler Architecture A, is not needed anymore, thus Architecture B can handle a broader range of disruptions.

A second example can be conceived concerning a more complex disruption, one involving the "common definition of the situation" (row 6 in Table 1) as would be handled by Architecture B. Let us assume that one agent asks a very personal question to another agent who is a stranger, such as "Has your wife fixed that problem?". This is recognized as a disruption by the enhanced disruption-recognition module EDR. The disruption is mapped by the MDRS module to a recovery strategy: in this case, an attempt by the agent to change the topic of the conversation. The recovery strategy is then set as a goal for the planner module PM that outputs a communicative act, namely an utterance about the current weather report that is executed by the execution module AE. The PM, however, reasons about the developing situated scenario to avoid proposing topics that can constitute a further disruption, for example discussing politics or religion. Differently from the simpler Architecture A, in Architecture B further disruptions can be prevented by planning. Note that the presence of a planning module in a social robot in the bartender scenario has been already proposed and studied by Foster and colleagues (2012); however, here the emphasis is on disruptions and consequent recovering.

Architectures A and B as presented above are proofs of concept to show how it would be feasible to realize and incorporate disruption recognition, recovery, and prevention in an artificial agent, at least in some cases. However, a more systematic approach is needed to define the requirements that an artificial agent should meet for realizing the recognition, recovery and prevention of the complete set of disruptions presented in Table 1. The first consideration is that such an agent should have spatial, conversational, representational (both general knowledge and knowledge on the current state of affairs), functional and reasoning capabilities. In fact, the disruptions presented in Section 4 and their corresponding recovery strategies mix different agent competences, as shown in the Requirements Table (Table 2) where we map the disruptions to requirements for the agent architecture. Table 2 shows how different disruptions could in principle be detected and recovered by only a subset of the additional capabilities, giving a useful guide for the implementation of the artificial agent. More complex agents (in terms of capabilities) could handle growing sets of interactional disruptions.

**TABLE 2. Requirement list for functional and social disruptions**

| FUNCTIONAL DISRUPTIONS | REQUIREMENTS | | | | PLANNING |
|---|---|---|---|---|---|
| | Agent Competence | Knowledge about | Current Representation | Reasoning | |
| An agent asks to change a previous Interaction | Conversational, Representational | Scenario | Current scenario Situated scenario | | Functional |
| An agent does not start the functional process | Representational Perceptual | Functional process | | | Functional |
| An agent does not proceed with the functional process | Representational | Functional process | Current scenario Situated scenario | | Functional |
| An agent interrupts the functional procedure | Representational | Functional procedure | Situated scenario | | Functional |
| Functional incompetence, performative mistake | Representational Functional, Reasoning | Functional goals | Situated scenario | | Functional recovery |
| SOCIAL DISRUPTIONS | | | | | |
| Disruption of the *common definition of the situation* | Representational | Scenario | Current scenario | | Recovery actions / illocutory acts |
| Disruption of *tacit norms* and related *expectations* | Representational, Reasoning | Tacit norms | Situated scenario | Expectations about tacit norms | Recovery illocutory acts |
| Disruption of *proxemic norms*: | Representational, perceptual, spatial | Proxemic norms | Current spatial status | Spatial | Spatial, recovery actions, recovery illocutory acts |
| Disruption of *conversational norms/rules* | Conversational, representational, perceptual | Conversational norms | Communication Status | Pragmatics | Recovery illocutory acts |
| Conversational "*repairables*": -Non-receptions -Misunderstandings -Speech errors | Conversational, perceptual, Self perception Self assessment | Illocutory acts | Communication Status, current illocutory acts | Self assessment on locutory acts. Assessment of perlocutory acts. | Illocutory acts |

| Disruption of the "*membrane*" | Conversational | Social norms, scenario | Current scenario | Social | Recovery actions/illocutory acts |
| Carry out *socially dispreferred actions* | Conversational, spatial | Social norms | | Social | Recovery actions/illocutory acts |

It is important to discuss the role of culture in the more complex agents that would be needed to fulfill the requirements listed in Table 2. In general, we can assume that the content would be culture-dependent whereas the representational, functional, reasoning, and planning capabilities do not need to be. In particular, the functional capabilities can be general, namely, to manage successfully the technical and physical intricacies of actually performing the duties of a bartender in our guiding example. These can be integrated with culture-specific content that, together with general representational reasoning and planning capabilities, smooths the interaction by detecting, recovering, and preventing interactional disruptions. A strong point of our proposal is, in fact, a sort of architectural decoupling between the functional and the expressive order that allows for a potential general application of the techniques of disruptions handling over a wide range of functionalities, tasks and contexts. To this end it is possible to use the guidelines for the practical application provided in Appendix A that also distinguish a multi-cultural context or situation, from a culturally-adaptive HRI application. A possible architecture that incorporates general cultural capabilities in a robot has been proposed by Borgo & Blanzieri (2019), where cultural knowledge, modeled as organized clusters of traits, interacts with planning and execution modules.

In order to highlight the practical feasibility of our proposal it is important to point out how some of the requirements of Table 2 are already realized, typically with slightly different goals, in robot architectures presented in the literature. For example, reasoning and conversational competences interact in the TeamTalk architecture (Marge and Rudnick, 2019) that manages miscommunication errors in human-robot dialogue. Miscommunication can be thought to correspond to the conversational "repairables" of our Table 2. In another example of relevant architecture, the rather complex relation between culture and sociality of agents is addressed by the Social Importance Dynamics (SID) model (Mascarenhas et al., 2013a) where cultural aspects are added to agents with perceptual, deliberative (reasoning in our terms), and planning competences. Such cultural flavors and social competence are important to manage what we called disruption of the "membrane", as well as the carrying-out of socially dispreferred actions in the last lines of Tables 1 and 2. Another relevant architecture is proposed in the CARESSES project (Khaliq et al., 2018), where culture-related information is considered and managed by keeping it separated and orthogonal to the other modules that realize the competences of the robot. This characteristic, that is also present in our Architecture B, is extremely important in practice, given the culturally-situated nature of interactional disruptions.

# 6. Discussion

The need to equip a robot with social skills and knowledge about (local) social norms is not new. For example, the compliance of a robot to social norms has been recently addressed in a real-world interaction (Gallo et al., 2021). The authors consider a robot and a human in a shared elevator scenario, with the artificial agent performing actions in the spectrum of machine-like and human-like behaviors to individuate the right mix of actions deemed human-friendly. To this end, the need of cross-domain social-norms knowledge that interacts with planning and execution module has been already proposed (Carlucci et al., 2015) and, considering the more general notion of agent, architectural proposal for having an agent that appears to be socially believable dates back at least two decades (Guye-Vuilleme & Thalmann, 2000). However, our approach, which focuses on interactional disruptions and their recovery and/or prevention, is original. The advantage is that we do not address the overall complexity of social interaction. Instead, we concentrate on some specific repairable issues that, as categories, are cross-cultural and cross-functional. In fact, multimodality of communication is not the only factor of complexity in human-machine interaction; ethnomethods, interactional norms (proxemics, talk) and the expressive dimension also play a role. Moreover, it is our stance that, particularly when casual interaction is concerned, we should go beyond user-personalized interactions (Onyeulo & Gandhi, 2020; Andriella et al., 2020). Adaptation should be especially tuned towards the developing phenomenal and social context (situated scenario) within a given cultural environment: we call these *context-personalized interactions* (knowing that such context is multilayered, cf. Sect. 1).

In order to collocate our theoretical proposal in the wide field of social robotics, it is useful to consider the dimensions of: appearance, social capabilities, purpose and application area, relational role, autonomy and intelligence, proximity, and temporal profile (Baraka et al. 2020) defined for characterizing the contributions in the research area. The dimensions that appear to be relevant for our approach are three: social capabilities (i), relational role (ii), and proximity (iii), whereas the others are not critical. *i)* In terms of social capability, we assume an artificial agent is capable of multimodal communication, i.e., natural language and non-verbal modalities. Moreover, it suffices to have a so-called "social interface" (one in which social behavior is modeled at the interface level), possibly lacking a deep model of social cognition. From this perspective, it is worth recalling that on casual interactional occasions, hence in the absence of closer/durable social relationships, the interaction order is "so fragile" that people monitor "any sign of disruption" (Tavory & Fine, 2020), ready with a recovery plan to enact upon need. It is especially on such casual (vs. intimate) occasions, that we may want artificial agents to operate. *ii)* The relational role the agent has, is to serve some utility on a given task (robot "for you" in Baraka et al.'s (2020) terms, bartending in our guiding example), but emulating a particular social trait found in humans (robot "as you", in this case the recovery strategies for interactional disruptions). *iii)* In the proximity dimension, the agent we are considering is co-located and physically interacting. Wrapping up, we aim to devise a relatively simple physical-conversational robot equipped with a relatively shallow social interface based on the handling of disruptions, including the interaction order and expressive layer, and the knowledge to represent the scenarios. Note that we are carefully limiting our target scenarios. For instance, more complex cases would require an artificial agent able to manage "disruptions-for". These are beyond the scope of this article and remain a challenge for future research.

There are some areas of agent and robot research that can be thought to be related to our approach. There is a rather long tradition of considering culture in conversational agents first reviewed by Rehm (2010), and it is interesting to note how politeness, namely having agents able to act politely across cultures, is considered from the outset. However, our emphasis is to manage interaction when things go wrong; that is hard to manage just with politeness per se as it asks for the kind of requirements at which we pointed in the previous section. Moreover, it is important to highlight that we do not address all the possible failures that can occur during human-robot interaction, just the ones that occur at the expressive level. The vast literature about failures reviewed by Honig and Oron-Gilad (2018) falls short, in our opinion, to address them, because it just considers the failures due to the robot, whereas here we also consider that the artificial agent could recover if equipped with suitable capabilities. Finally, we have not addressed the issue of social presence in computer mediated communication reviewed by Oh and colleagues (2018). However, it is interesting to notice that what emerges from such literature is that being "exposed to cues that indicate a social context

(e.g., conversation, partner, group, etc.) can lead to heightened levels of social presence". Maintaining and preserving the social context from possible disruptions when interacting with artificial agents is our goal here, and so it would be possible to argue that we expect that the construct of social presence could be detected once the necessary requirements are fulfilled. On the other hand, it is social presence, as precondition, that triggers the social phenomena, i.e. the intervening of the expressive level and its possible disruptions, that we have addressed in this paper.

It is worth emphasizing once more how our approach based on representing, detecting, and recovering disruptions has deep roots in sociological research and is consistent with recent advancements (Tavory & Fine, 2020). The artificial agent we propose may have a shallow social interface, but one of the important points to the advantage of our proposal is that, also between humans, interaction happens at the interfaces, and the disruptions of norms, membranes, shared situations and so on, that we listed in Table 1 coupled with recovery strategies, are the normal way in which people interact. Although we focused on modular architectures, we argue that the requirements we have identified, could be also realized by Machine Learning, provided that the proper tasks are defined, data collected, or, in the case of Reinforcement Learning, reward functions defined. Equipping a robot with the capability that we described can go a long way towards the realizations of effective social human-robot interactions.

# 7. Conclusions

In this paper, we presented the idea that, by handling the interactional disruptions with specific recovery strategies, an artificial agent can smooth the interaction with humans, and can do so in a culturally-adaptive way. With the help of a simple but illustrative guiding example, and levering on microsociological literature, we summarized the main types of disruption that can occur in human-human interaction. This classificatory work allows artificial agents engaged in human-robot interaction to *reason about, and in case contribute to, the recovery from disruptions of the "interaction order" and its expressive dimension*. We then illustrated how different agent architectures can be devised in order to integrate these social skills in artificial agents. Finally, we proposed general requirements for such agents.

Our main contribution is to connect existing literature on interactional disruptions to the realm of artificial agents in a way that is both compact and translatable into practical requirements. Our examples of architectures and our map of the disruptions to the requirements show that realizing an artificial agent that successfully handles the nuances of social disruptions is in principle possible. We would like to emphasize that the considered disruptions are general and possibly occur in any task that requires human-robot interactions. The proposed architectures and requirements treat disruptions as general and as possibly occurring in a wide range of tasks. Moreover, the dependency on culture is maintained to a minimum and included in some modules in the Architecture A and, when considering the general requirements, confined to the content of the representations in terms of knowledge and current state of affairs. In this way, an artificial agent with a specific functionality can also have a general competence on the expressive order that, with culture-specific ways to properly handle the disruptions, allows to deploy instances of the agent in different cultures. Finally, the competence on the expressive order could be used also for different functionalities, and consequently the effort to build culture-specific content to be represented for handling the disruptions can be exploited for a wide range of practical functionalities.

The cultural nature of the approach allows also to modulate the kind of interaction the artificial agent's designer would like to instantiate to handle interactional disruptions. Although the default of our approach appears to be that the artificial agent should act in a socially-smooth way as a human would do, and although there is evidence that people "tend to prefer robots better complying with the social norms of their own culture" (Bruno et al., 2019), it is anyway advisable to further consider, for each specific application, the cultural variation concerning humans' conceptions of artificial agents and dispositions towards interacting with them (e.g. Evers et al., 2008; Nomura et al., 2008; Lim et al., 2021). For example, different cultures entail different expectations about the behavior of robots in terms of proxemics, also with respect to what is expected to be the proxemic conduct of humans in that cultural context (e.g., Eresha et al., 2013; Joosse et al., 2014). Moreover, and possibly more importantly, it should be kept in mind that people regard robots as not fully competent social members (Pitsch, 2016), as it happens, for instance, with animals, children and people labeled as "mentally ill". This qualifies the way humans interact with artificial agents, and this too varies culturally. Obviously, the whole complexity of culture-dependent behaviors goes beyond todays' possibilities. However, the focus on interactional disruptions and their recovery tackles a new and critical part of the human-robot interaction, one that can provide a proper advantage in terms of smoothness and quality of the interaction, even with relatively simple architectural solutions.

## ACKNOWLEDGEMENT

The authors are listed in alphabetical order. CB, EB and SB conceived the research. SM defined the guiding example. CB, SB and SM redacted the disruptions table. EB and SB individuated the architectures and the requirements. CB, EB and SB elaborated the guidelines. All the authors searched the literature, discussed the intermediated results, and contributed to drafting, editing and revising the manuscript.

**Appendix A**

These guidelines, in the form of Q&A, are provided to help the practitioner to include (in full or in part) in a given agent architecture the requirements we proposed. Note that culture, as intended below, is a broad term that includes, but does not coincide with, national heritage; for example, organizations or groups of professionals develop rather specific cultures. Moreover, to avoid the risk of the definition and application of stereotypes, culture should NOT be conceived just as a property of the user, but rather as a property of the ensemble of participants and the activity in which they are involved.

1. Is your agent/robot (also) a social agent/robot?
   YES Chances are that the interaction with humans, for example in a collaborative setting, will also involve the expressive order (as mentioned in Section 3) and our proposal directly applies.
   NO Interaction with the agent/robot, if any, has been analyzed and designed in terms of other HRI concepts, for example ergonomics or collision avoidance; however, social interaction phenomena that triggers the expressive order could still occur.
2. Have you experienced failures of your social agent/robot that can be considered to be at the expressive order?
   YES Such disruptions, as listed in Table 1, are common between humans and we propose that their detection and recovery could be integrated in the agent/robot architecture leveraging the data about the failures.
   NO Pay anyway attention to human failures and recoveries of the expressive order, as failures could be also on the human side and the agent could help to recover them, Table 1 should help to check.
3. Do you have one or more social contexts, like the bartender scenario or others, that you expect to be relevant in the deployment of the agent/robot?
   YES The expressive order is definitely relevant: on the one, hand the social context could be a useful guide to detect disruptive behavior; on the other hand, the recovery strategies (see Table 1) could help to maintain the context itself.
   NO Pay anyway attention to the fact that your agent/robot could be immersed in a context that triggers the expressive order (see 4. and 5. below).
4. Does your agent/robot participate in co-present multi-party human interaction?
   YES Even if your agent/robot is not social, the presence of more than one human generates a social context, i.e. humans that interact together with an agent/robot, and that implies the expressive order and its possible disruptions.
   NO Pay anyway attention to the fact that your agent/robot in interaction with a single user, could be immersed in a context involving other humans that hence triggers the expressive order.
5. Is the achievement of social presence a goal of your project?
   YES Social presence can be facilitated by managing the expressive order, and conversely social presence triggers the expressive order so our proposal is extremely relevant in particular for single-user interaction.
   NO If the expressive order is somehow triggered there is anyhow the need to manage disruptions and recoveries to smooth the interaction.
6. Which competences, as listed in Table 2, does your agent/robot have? (perceptual, conversational, representational, functional, reasoning, spatial, self perception, self assessment) If the interaction triggers the expressive order, check in Table 2 which disruption it is possible to manage with the competences already available in your architecture.
7. Do you have a planning module in your architecture?
YES If the interaction triggers the expressive order, the planning module can be used to recover the disruptions as in Architecture B
NO You can use a reactive approach as depicted in Architecture A.
8. Is your architecture culturally-adaptive?
YES We expect that social contexts are culture-dependent, you can take advantage of your architecture in order to deal with them and their disruptions of the expressive order devising an architecture more complex of our simple Architectures A and B.
NO Disruptions are quasi-universal whereas the specific ways of recovering them are not, you could introduce some culture-adaptiveness just to deal with them using, for example, one of our architectures.
9. Is your agent/robot going to be deployed in different cultural contexts?
YES You should expect the need of different ways to recover from the disruptions of the expressive order depending on the culture, and include in your architecture some culture-adaptiveness (see above Q8) that can be statically set for each specific deployment.
NO In presence of just one cultural context the development of the disruption recovering part is simplified (for example the database in Architectures A and B does not need to be multicultural).
10. Are cultural differences relevant for the activity of your agent/robot?
YES For culture-sensitive activities (for example the ones related to the body where proximity and contact play a role) we expect that the architecture should also be statically culture-adaptive (see 8. above); if not, adding the culture-dependent recovery strategies for the disruption at the expressive order (Table 1) could be beneficial.
NO In this case, as 9. above, the development of the recovery part is simplified.
11. Are cultural differences among the present humans relevant?
YES In this case you have a multi-party interaction with cultural differences and your architecture needs to be dynamically culturally-adaptive (beware of the pitfall of stereotypes on detecting the culture of a user), and the expressive order is certainly relevant; please note that our simple Architectures A and B do NOT cover this case.
NO Also in this case, as above 9. and 10. above, the development of the recovery part is simplified.
12. Is your context multicultural in itself?
YES If cultural differences are relevant (see 10. above), we expect dynamic culture-adaptiveness, as 11. above, or the assumption of a general international/globalized social context (for example in an airport) for dealing with the expressive order.
NO In this case, you can treat the differences in culture as incidental and the disruption recovery strategies in a multicultural version could be beneficial (beware of the pitfall of stereotypes on detecting the culture of the user) although they are not covered by our Architectures A and B.
13. Should your agent/robot detect culture-based misunderstandings?
YES Given the culture-dependent nature of the actuation of the recovery strategies, our framework could be useful to detect misunderstandings due to different ways of dealing with the expressive order; however, this is not covered by our Architectures A and B.

NO In this case the disruptions at the expressive order are simplified and can be dealt with by our approach.
14. Is your agent/robot a cultural mediator?
YES This the most complex case, let us observe that it is unavoidable to deal somehow with the expressive order and our Architectures A and B do NOT cover this.
NO As in 13. above, the management of the expressive order is simplified and it can be dealt with by our approach.